\documentclass[11pt, a4paper]{article}

\usepackage{amsmath,amssymb,graphicx}

\usepackage{authblk}
\usepackage{faktor}
\usepackage{caption}
\usepackage{subcaption}
\usepackage[section]{placeins} 
\usepackage{float}
\usepackage{listings}
\usepackage{color}

\usepackage[round]{natbib}

\usepackage[draft]{hyperref}

\usepackage{times}

\graphicspath{{.}{./Figures/}{./Figures_2Dfunction/}{./Figures_1Dfunction/}{./Figures_Neural_Networks/}{./Figures_LinearRegression/}{../notes/Images/}{../notes/}}

\newlength{\widthadd}  
\newlength{\heightadd}  
\begin{document}

\title{\sffamily Gradient descent with momentum --- to accelerate or to super-accelerate?}

\date{}


\author[1]{Goran Nakerst}
\author[1]{John Brennan}
\author[1,2]{Masudul Haque}

\affil[1]{\small Department of Theoretical Physics, Maynooth University, Co.\ Kildare, Ireland}
\affil[2]{\small Max-Planck Institute for the Physics of Complex Systems, Dresden, Germany}



\maketitle

\begin{abstract}
We consider gradient descent with `momentum', a widely used method for loss function minimization in
machine learning.  This method is often used with `Nesterov acceleration', meaning that the gradient
is evaluated not at the current position in parameter space, but at the estimated position after one
step.  In this work, we show that the algorithm can be improved by extending this `acceleration' ---
by using the gradient at an estimated position several steps ahead rather than just one step ahead.
How far one looks ahead in this `super-acceleration' algorithm is determined by a new
hyperparameter.  Considering a one-parameter quadratic loss function, the optimal value of the
super-acceleration can be exactly calculated and analytically estimated.  We show explicitly that
super-accelerating the momentum algorithm is beneficial, not only for this idealized problem, but
also for several synthetic loss landscapes and for the MNIST classification task with neural
networks.  Super-acceleration is also easy to incorporate into adaptive algorithms like RMSProp or
Adam, and is shown to improve these algorithms.

\end{abstract}

\hrule

\bigskip \bigskip \bigskip 


\section{Introduction and Main Result}

Training machine learning models very often involves minimizing a scalar function of a large number
of real parameters.  This function is usually called the loss function or cost function in machine
learning, and often called the objective function in the optimization literature.
Arguably, the most popular class of methods for minimization are first-order methods based on
gradient descent \citep{LeCun_Bengio_Hinton_Deeplearning_Nature2015, Ruder_overview_2016,
  book_Goodfellow_Bengio_deeplearning_2016}.  For example, for training (deep) neural networks,
these are the preferred class of methods because gradients of the loss function with respect to
parameters can be easily calculated via backpropagation \citep{Rumelhart_Hinton_backprop_Nature1986,
  LeCun_Bottou_Mueller_efficientbackprop_1998, book_Nielsen_NeuralNetworks_2015}, whereas
calculating second derivatives or the Hessian is expensive.

The  algorithm for basic gradient descent is the iteration 
\begin{equation}  \label{eq:pure_gd} 
	\theta^{(i+1)} =\theta^{(i)} - \eta \nabla L(\theta^{(i)}),
\end{equation}
Here $\theta$ is a $D$-dimensional vector, consisting of the parameters of the machine learning
model.  The superscripts represent the iteration count.  The objective function to be minimized,
$L(x)$, is a $\mathbb{R}^D\to\mathbb{R}$ function known as the loss function or cost function.  The
positive constant $\eta$ is the learning rate.  It is often modified during minimization, i.e.,
could depend on the iteration count $i$.  In this work, we will mainly focus on constant $\eta$.

We consider a commonly used variation of gradient descent:
\begin{equation}  \label{eq:plain_momentum} 
 \begin{split}  
  m^{(i)} &= g m^{(i-1)} - \eta \nabla L(\theta^{(i)}) \\
  \theta^{(i+1)} &= \theta^{(i)}  + m^{(i)} .
 \end{split}
\end{equation}
This is generally known as gradient descent with momentum, and also as Polyak's heavy ball method
\citep{Polyak_1964}.  The direction of descent at each step is a linear combination of the direction
of the gradient and the direction of the previous step.  The hyperparameter $g$ provides an
`intertia'.  Although some suggestions for varying $g$ can be found in the literature
\citep{Srinivasan_Sankar_ADINE_2018, Chen_Kyrillidis_decaying_momentum_2019}, the value $g=0.9$ is
so widely used that we will mostly focus on this value unless explicitly stated.  Momentum is often
used by evaluating the gradient not at the current position in parameter space, but rather at the
estimated position at the next iteration:
\begin{equation} \label{eq:usual_Nesterov}
 \begin{split}  
   m^{(i)} &= g m^{(i-1)} - \eta \nabla L(\theta^{(i)}+ g m^{(i-1)})  \\
   \theta^{(i+1)} &= \theta^{(i)}  + m^{(i)} .
 \end{split}
\end{equation}
This modification is commonly known as Nesterov's accelerated momentum
\citep{Nesterov_1983}.  The original algorithm of
\citet{Nesterov_1983} looks somewhat different, but the form
\eqref{eq:usual_Nesterov}, introduced in
\citet{Sutskever_Hinton_momentum_2013}, is what is more commonly known
in current machine learning literature as Nesterov acceleration.  The
words `momentum' and `acceleration' are both used in a different sense
from their meaning in physics/mechanics.

In this paper, we propose and analyze the  following modification to the algorithm
\eqref{eq:usual_Nesterov}:
\begin{equation} \label{eq:super_Nesterov}
\begin{split} 
 m^{(i)} &= g m^{(i-1)} - \eta \nabla L(\theta^{(i)}+ \sigma m^{(i-1)})  \\
 \theta^{(i+1)} &= \theta^{(i)}  + m^{(i)} .
\end{split}
\end{equation}
We have introduced a new hyperparameter, $\sigma$.  When $\sigma=0$, one obtains the heavy-ball
method \eqref{eq:plain_momentum}, and when $\sigma=g$, one obtains Nesterov's usual acceleration,
\eqref{eq:usual_Nesterov}.  In this work we show that it can be advantageous to use values of
$\sigma$ significantly larger than $g\approx1$.  Instead of using the gradient at an estimated point
one step ahead, this is using the gradient at an estimated point \emph{several} steps ahead.  As
this is an extension/strengthening of Nesterov's ``acceleration'' idea, we refer to $\sigma\gtrsim1$
as ``super-acceleration''.

\begin{figure*}
\begin{center}
\includegraphics[width=1.0\textwidth]{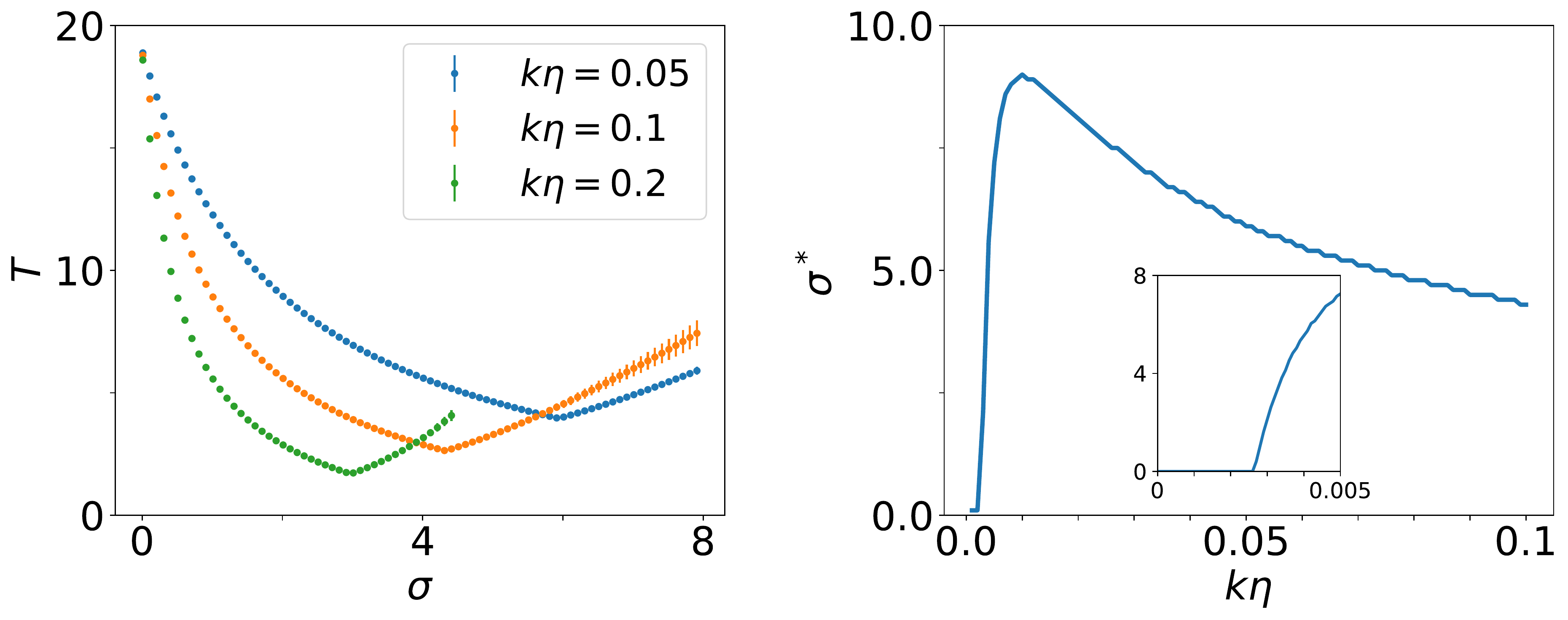}
\caption{The advantage of super-accelerating the Nesterov momentum
  algorithm, for minimization of the parabolic objective function
  $L(\theta)=\frac{1}{2}k{\theta}^2$ of a single parameter $\theta$.  (a)
  The time scale $T$ for reaching the minimum of $L(\theta)$ is a
  non-monotonic function of the super-acceleration parameter $\sigma$,
  and has a well-defined minimum.
  (b) The optimal value of $\sigma$, i.e., the value of $\sigma$ for
  which $T$ is minimal is plotted as a function of $k\eta$, where
  $\eta$ is the learning rate.  Except for very small and very large
  $k\eta$, the optimal value is seen to be significantly larger than
  $1$.
\label{fig:introplot_1Dparabolic}
  }
\end{center}
\end{figure*}

We will first examine the benefit of super-acceleration in the `toy' setting for convex
optimization: a parabolic loss function in a one-parameter system, $L(\theta)=\frac{1}{2}k\theta^2$.
The detailed analysis is in Section \ref{sec:1D_quadratic}, but here we provide an summary of the
main lessons.  The algorithm \eqref{eq:super_Nesterov} moves the parameters exponentially toward the
minimum (possibly with oscillatory factors).  The decay constant in the exponent provides the time
scale $T$ for minimization.  In Figure \ref{fig:introplot_1Dparabolic} we show the value of the
super-accelearation parameter $\sigma$ that minimizes this time scale.  The optimal value $\sigma^*$
of $\sigma$ depends on the learning rate $\eta$ in the combination $k\eta$.  Except for extremely
small $k\eta$, the optimal $\sigma$ is generally significantly larger than $1$.  This provides basic
support for our proposal that super-accelerating Nesterov momentum (using $\sigma\gtrsim1$, i.e.,
looking ahead several estimated steps) may be advantageous in training by gradient descent.  Of
course, the generalization from a quadratic one-dimensional problem to a generic machine-learning
problem is highly nontrivial.  We therefore tested the effect of $\sigma$ for two-dimensional and
$\approx50$-dimensional minimization problems (Section \ref{sec:high_D}), and for training neural
networks for handwritten digit classification on the MNIST dataset (Section
\ref{sec:NN_MNIST_classification}).  In all cases, we find that extending $\sigma$ beyond $1$
provides significant speedup in the optimization/training process.

In addition to momentum and acceleration defined above, many other variants and improvements of
gradient descent are in use or under discussion in the machine learning community.  Listings and
discussions of the most common algorithms can be found in \citep{Ruder_overview_2016,
  Mehta_Bukov_ML_for_physicists_2019, Sun_Cao_survey_methods_2019, Zhang_optimization_review_2019}.
Overviews are also provided in Chapter 8 of \citet{book_Goodfellow_Bengio_deeplearning_2016}, and
Chapter 5 of \citet{book_Kochenderfer_Wheeler_algorithms_2019}.  Specially prominent are methods
that adaptively modify the learning rate $\eta$, such as AdaGrad
\citep{Duchi_Hazan_adaptive_subgradient_2011}, RMSProp \citep{rmsprop_Courseralecture_2012}, Adam
\citep{Kingma_Ba_adam_2014} and Adadelta \citep{Zeiler_adadelta_2012}.
Momentum and acceleration can be readily incorporated into these adaptive schemes
\citep{Dozat_Nadam_2016, Ruder_overview_2016, book_Goodfellow_Bengio_deeplearning_2016}.  Hence,
super-acceleration (arbitrary $\sigma$) can also be defined.  The analysis of super-accelerating one
of the modern adaptive algorithms is significantly complicated.  However, we provide a preliminary
analysis for RMSProp (Section \ref{sec:RMSPROP}) which shows that using $\sigma>1$ should benefit
RMSProp as well.

This Article is organized as follows.  
In Section \ref{sec:1D_quadratic}, we provide a detailed treatment of the one-parameter parabolic
problem.  The exponential relaxation can be understood by approximating the algorithm by an ordinary
differential equation (ODE) which describes the damped harmonic oscillator.  Optimizing the
algorithm is then seen to be equivalent to tuning parameters to ``critical damping''; this allows us
to predict the optimal super-acceleration, as presented in Figure \ref{fig:introplot_1Dparabolic}.
In Section \ref{sec:high_D} we explore our algorithm in two higher-dimensional cases: a synthetic
non-quadratic objective function and a 51-dimensional quadratic objective function resulting from
linear regression.  This exploration shows that super-acceleration is broadly advantageous, but also
reveals possible interesting side-effects such as non-linear instabilities.  
In Section \ref{sec:NN_MNIST_classification} the algorithm is applied to the MNIST digit
classification problem.  Super-acceleration is applied to examples using both non-stochastic and
stochastic gradient descent.  
In Section \ref{sec:RMSPROP} we show how super-acceleration is readily incorporated into RMSProp,
and the benefits of this combination.
Finally, Section \ref{sec:concl} provides context and discussion.  


\section{Quadratic loss function for one parameter --- damped harmonic oscillators and optimal super-acceleration}
\label{sec:1D_quadratic}

In this section we analyze the effect of super-acceleration on the
minmization of a quadratic objective function of a single parameter $\theta$.  In the Introduction  
(Figure \ref{fig:introplot_1Dparabolic}) we have provided an
`executive summary'; here we consider the problem in some detail.  

We are interested in gradient descent with super-acceleration applied to minimizing the loss
function of the form $L(\theta)= \frac{1}{2}k\theta^2$.  Here $\theta$ is a single real variable,
and $k$ is a positive constant.  This is the simplest and most natural model potential for a convex
function of a single variable.  The algorithm for this special case is
\begin{equation}  \label{eq:super_Nesterov_1D_a}
 \begin{split}  
   m^{(i)} &= gm^{(i-1)} - k \eta  (\theta^{(i)} + \sigma m^{(i-1)})  \\
   \theta^{(i+1)} &= \theta^{(i)}  + m^{(i)} ,
 \end{split}
\end{equation}
which can be rewritten as 
\begin{equation}  \label{eq:super_Nesterov_1D_b}
 \begin{split}  
	m^{(i)} - m^{(i-1)} &= \alpha m^{(i-1)}- k \eta \theta^{(i)}  \\
	\theta^{(i+1)} - \theta^{(i)}  &=  m^{(i)} , 
 \end{split}
\end{equation}
with $\alpha=g-k\eta\sigma-1$ a negative real number.  Written in this form, the iteration equations
are reminiscent of difference equaions that are obtained upon discretizing ordinary differential
equations (ODEs), as commonly done in numerical solutions of ODEs \citep{book_Golub_Ortega_1992,
  book_Stoer_Bulirsch_2002}.  It turns out that equations \eqref{eq:super_Nesterov_1D_b} can be
obtained as approximations to a differential equation representing damped harmonic motion
\citep{Ning_momentum_1999, Su_Boyd_Candes_diffeq_for_Nesterov_accel_2014,
  Kovachki_Stuart_analysis_momentum_2019}.  In fact, there are multiple differential equations
which, upon discretization, can lead to the scheme of Eq.\ \eqref{eq:super_Nesterov_1D_b}, depending
on whether the derivatives are approximated by a forward differnce rule, a backward difference rule,
or a midpoint rule.

In Subsection \ref{subsec:timescale}, we explain how the representation in terms of a damped
harmonic oscillator allows us to define a time scale for the optimization process and thus leads to
the prediction of optimal values of $\sigma$, as presented in the Introduction.
In Subsection \ref{subsec:ODEs}, we will outline several possible ODEs and compare how their
predictions match the behavior of the algorithm.
Subsection \ref{subsec:optimal_1D} discusses the optimal values of the
super-acceleration predicted by this analysis.

\subsection{Time scale for minimization --- critical damping}
\label{subsec:timescale}

We will argue below that the algorithm \eqref{eq:super_Nesterov_1D_b} minimizes the variable
$\theta$ exponentially, with possible oscillations, i.e., the evolution of $\theta$ is
well-described by the form $e^{-t/T}\cos(\bar{\omega}t+\delta)$, where $t$ is a continuous
counterpart of the iteration count $i$.  When the dynamics is of this form, the quantity $T$ is the
``timescale'' for the minimization process, i.e., the inverse of the rate at which the algorithm
approaches the minimum $\theta=0$.  The goal of optimizing the algorithm is thus to choose
hyperparameters that will minimize the timescale $T$.  Other criteria, such as minimizing the time
it takes to come within a small distance $\epsilon$ from $\theta=0$, is often used in the more
mathematical literature on optimization.  We consider the smallness of the relaxation timescale $T$
to be a more practically relevant and physical measure of the performance of an optimization
algorithm.

The algorithm \eqref{eq:super_Nesterov_1D_b} is approximated by a differential equation of the form 
\begin{equation}
\label{eq:damped_harmonic_oscillator_0}
\frac{d^2x}{dt^2} + A \frac{dx}{dt} + B x = 0,
\end{equation}
with $B>0$ and $A>0$.  Here the function $x(t)$ is a continuous
version of  $\theta^{(i)}$.  Expressions for $A$ and $B$ in terms of
$\alpha$ and $k\eta$ are discussed in the next subsection.  Here, we
consider the consequence of this form of differential equation.

Eq.\ \eqref{eq:damped_harmonic_oscillator_0} describes damped harmonic oscillation, well-known from
introductory classical mechanics \citep{LandauLifshitz_mechanics_1982,
  book_Greiner_classicalmechanics_2003, book_Thornton_Marion_ClassicalDynamics_2004}.  The solution
has the form
\begin{equation} \label{eq:dampedHO_soln_1}
x(t) = e^{-tA/2} \left[ C e^{i\omega t} + D e^{-i \omega t} \right],
\end{equation}
with $\omega=\sqrt{k\eta -A^2/4}$.  For small $A$ (`underdamped' regime), the solution is
oscillatory with an exponentially decreasing envelop.  In this regime the relaxation constant can be
read off from the exponential factor to be $T=2/A$, which decreases with $A$.  

For $A>\sqrt{4B}$ (`overdamped' regime), the oscillation frequency $\omega$ becomes imaginary
($\omega=i\omega''$), so that the dynamics is non-oscillatory.  Now, the dynamics consists of two
exponentially decaying terms, with rate constants $A/2\pm\omega''$.  There are different timescales
corresponding to the initial and the long-time relaxation.  One may interpret either of
$A/2\pm\omega''$ as $1/T$.  In either case, $T$ increases with $A$ in the overdamped regime.  (When
we plot $T$ in Figure \ref{fig:introplot_1Dparabolic}, we use the initial relaxation timescale,
because this seems more relevant to the efficiency of training a machine learning model.)

\begin{figure*}
\begin{center}
\includegraphics[width=1.0\textwidth]{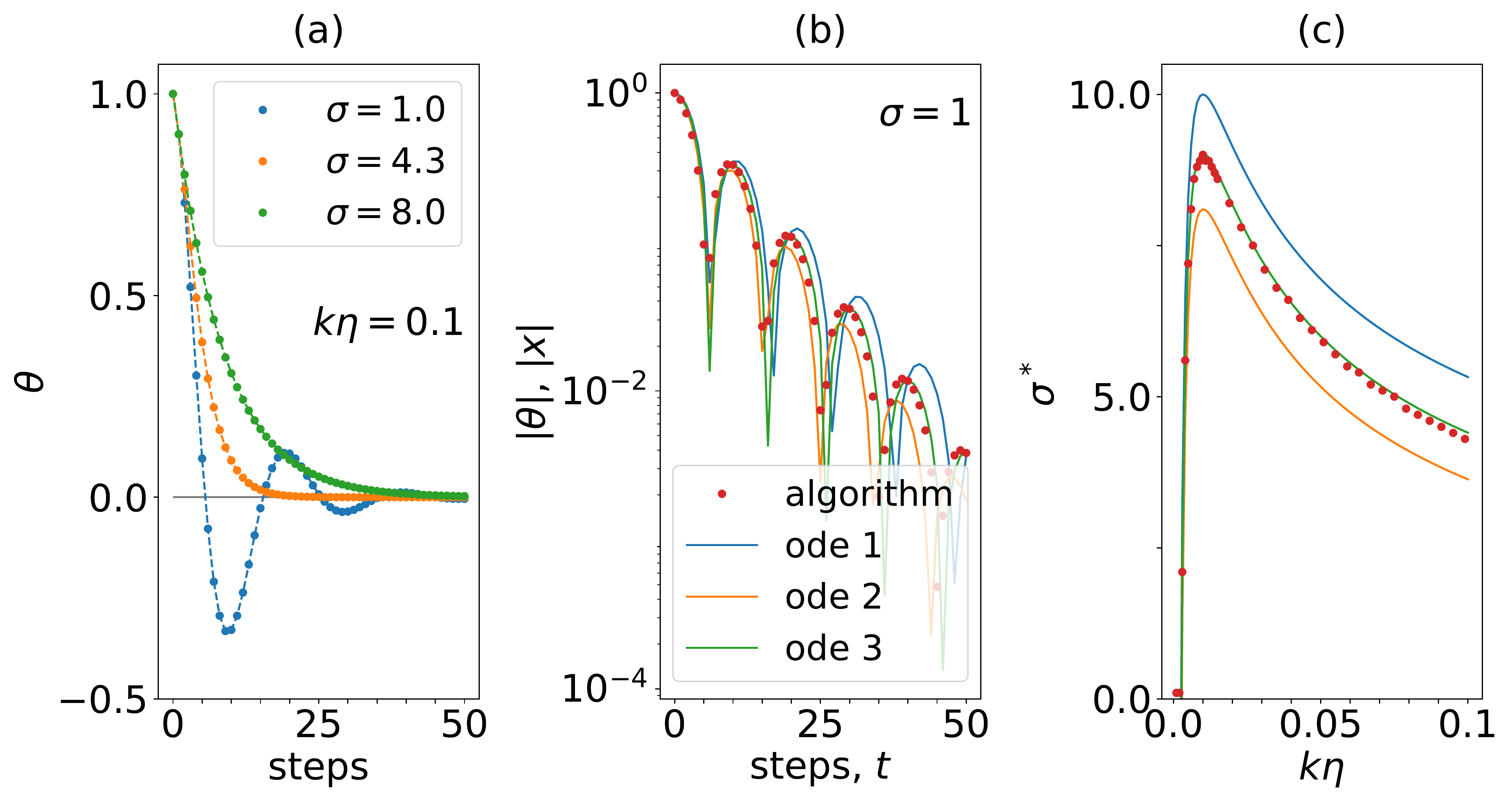}
\caption{(a) The points show the $\theta^{(i)}$ values followed by the algorithm for $k\eta=0.01$,
  for various super-acceleration values $\sigma$.  We show an underdamped, an overdamped and a
  critical case.  Each trajectory is fitted by an oscillatory exponental (dashed lines).  (b) For
  $k\eta=0.01$ and $\sigma=1$, we compare $\theta^{(i)}$ with the predictions for $x(t)$ from the
  three ODE's, Eqs.\ \eqref{eq:damped_harmonic_oscillator_1}
  \eqref{eq:damped_harmonic_oscillator_2}, \eqref{eq:damped_harmonic_oscillator_3}, labeled as ode
  1, 2 and 3 respectively.
(c) The value of $\sigma$ at which the relaxation is fastest, compared with predictions for the
  critical value obtained from the three ODE's.
\label{fig:1Dparabolic_fit_to_damped_cosine}
}
\end{center}
\end{figure*}

In Figure \ref{fig:1Dparabolic_fit_to_damped_cosine}(a), we show some example trajectories of the
algorithm \eqref{eq:super_Nesterov_1D_a} starting from $\theta=1$, together with fits to the
function $e^{-t/T}\cos(\bar{\omega}t+\delta)$, with fit parameters $T$, $\bar{\omega}$ and $\delta$.
The fits are seen to be excellent, demonstrating that the minimization process of algorithm
\eqref{eq:super_Nesterov_1D_a} is well-described by exponential relaxation with or without
oscillations, as predicted by the ODE approximation \eqref{eq:damped_harmonic_oscillator_0}.

As the relaxation constant $T$ decreases in the underdamped regime and increases in the overdamped
regime, the point of `critical' damping ($\omega=0$ or $A=\sqrt{4B}$) is where  $T$ is smallest.
It turns out that $A$ is a monotonically increasing function of $\sigma$ (Subsection
\ref{subsec:ODEs}).  Thus there is an optimal value, $\sigma=\sigma^*$, at which $T$ is a minimum.
This was the result shown and discussed in Figure \ref{fig:introplot_1Dparabolic}. In Figure
\ref{fig:introplot_1Dparabolic} we obtained the optimization timescale $T$ simply by fitting data
from the algorithm to the form $e^{-\theta/T}\cos(\bar{\omega}t+\delta)$.  Approximate analytical
expressions for $\sigma^*$ can be obtained from the differential equations, by finding the parameter
values corresponding to critical damping (Subsection \ref{subsec:optimal_1D}).

\subsection{Damped harmonic oscillators: ODEs approximating Eq.\ \eqref{eq:super_Nesterov_1D_b}}
\label{subsec:ODEs}

In the numerical solution of ordinary differential equations describing evolution, say Newton's
equation for the motion of a particle, it is common to distretize time into small steps and
represent derivatives as finite differences \citep{book_Golub_Ortega_1992,
  book_Stoer_Bulirsch_2002}.  Here, we reverse the approximation procedure: our starting point, the
algorithm \eqref{eq:super_Nesterov_1D_a} or \eqref{eq:super_Nesterov_1D_b}, is a finite difference
equation which we relate approximately to an ODE.  The procedure is not unique and several slightly
different ODEs can serve this purpose.  Fortunately, they all have the same form
\eqref{eq:damped_harmonic_oscillator_0}.

In Eq.\ \eqref{eq:super_Nesterov_1D_b}, we interpret $\theta^{(i)}$ as the discretized values of a
position variable $x(t)$.  The $i$'s are then integer values of $t$.  The second line of Eq.\
\eqref{eq:super_Nesterov_1D_b} then suggests that $m_i$ should be interpreted as the discretized
values of velocity, $v=\frac{dx}{dt}$, because the left side of this equation,
$\theta^{(i+1)}-\theta^{(i)}$, is the first-order forward difference formula for the derivative of
$x$, with step size unity.  In the first line of Eq.\ \eqref{eq:super_Nesterov_1D_b}, we could
identify the difference $m^{(i)}-m^{(i-1)}$ as the acceleration $\frac{d^2x}{dt^2}$.  We thus have
the ODE
\begin{equation}\label{eq:damped_harmonic_oscillator_1}
 \frac{d^2x}{dt^2} - \alpha \frac{dx}{dt} + k\eta x = 0.  
\end{equation}
An inconsistency in this procedure is that both $m^{(i)}$ and $m^{(i-1)}$ have been separately
identified as $v(t)$.  However, the error due to this inconsistency is of the same order as the
difference formulas.  If one insists on identifying $m^{(i)}$ as $v(t)$ also in the first line of
Eq.\ \eqref{eq:super_Nesterov_1D_b}, one obtains by using the backward difference for the derivative
of $v(t)$:
\begin{equation}\label{eq:damped_harmonic_oscillator_2}
\frac{d^2x}{dt^2} - \frac{\alpha}{1+\alpha} \frac{dx}{dt} + \frac{k\eta}{1+\alpha} x = 0,
\end{equation}
Another approach would be to identify $\frac{1}{2}\left[m^{(i)}+m^{(i-1)}\right]$ as $v(t)$, so that
the difference formula used for $v'(t)$ is now a second-order approximation.  This leads to
\begin{equation}\label{eq:damped_harmonic_oscillator_3}
\frac{d^2x}{dt^2} - \frac{\alpha}{1+\alpha/2} \frac{dx}{dt} + \frac{k\eta}{1+\alpha/2} x = 0.
\end{equation}
Since the difference formula for $x'(t)$ is still first-order, the relation between this last
equation \eqref{eq:damped_harmonic_oscillator_3} and the original discrete algorithm is still a
first-order approximation.  

In Figure \ref{fig:1Dparabolic_fit_to_damped_cosine}(b) we compare the oscillatory minimization
dynamics of the algorithm \eqref{eq:super_Nesterov_1D_a} or \eqref{eq:super_Nesterov_1D_b} to
numerically exact solutions of the three ODEs.  It is clear that Eq.\
\eqref{eq:damped_harmonic_oscillator_3} performs best as an approximation.

Although it is not of higher order than the other two ODEs, the empirical observation is that Eq.\
\eqref{eq:damped_harmonic_oscillator_3} reproduces features of the discrete algorithm (such as the
value of $\sigma^*$) more faithfully than the other two ODEs,
\eqref{eq:damped_harmonic_oscillator_1} and \eqref{eq:damped_harmonic_oscillator_2}.  This is
presumably because Eq.\ \eqref{eq:damped_harmonic_oscillator_3} incorporates a second-order
discretization approximation.

Of course, there may be additional ODEs which approximate the same finite difference equations, but
we confine ourselves to these three.   Our purpose is not to explore the space of possible ODE
approximations, but rather to show that all reasonable approximations are of the form
\eqref{eq:damped_harmonic_oscillator_0}, i.e., describe damped harmonic motion.

\subsection{Optimal parameter $\sigma^*$}  \label{subsec:optimal_1D}

The three ODE's, \eqref{eq:damped_harmonic_oscillator_1}, \eqref{eq:damped_harmonic_oscillator_2},
and \eqref{eq:damped_harmonic_oscillator_3}, each have the form
\eqref{eq:damped_harmonic_oscillator_0} discussed in Subsection \ref{subsec:timescale}.  Each of
these equations describe damped harmonic oscillations.  In each case, $A$ and $B$ can be expressed
in terms of $k\eta$ and $\alpha=g-k\eta\sigma-1$.  The case of critical damping is obtained from the
condition $A=\sqrt{4B}$.  Solving this condition for $\sigma$ gives estimates for the optimal
super-acceleration:
\begin{equation}\label{eq:nu_eta_critical_1}
\sigma^* = \frac{2}{\sqrt{k\eta}} - \frac{1-g}{k\eta},
\end{equation}
\begin{equation}\label{eq:nu_eta_critical_2}
\sigma^* = -2 + 2\sqrt{1 + \frac{1}{k\eta}} - \frac{1-g}{k\eta}
\end{equation}
and
\begin{equation}\label{eq:nu_eta_critical_3}
\sigma^* = -1 + \sqrt{1+\frac{4}{k\eta}} - \frac{1-g}{k\eta}.
\end{equation}
We can obtain the optimal super-acceleration parameter $\sigma^*$ directly from the discrete
algorithm, by fitting a function of the form $e^{-t/T}\cos(\bar{\omega}t+\delta)$ to the
$\theta^{(i)}$ vs $i$ data (identfying $i$ with time), and then finding the value of $\sigma$ for
which the timescale $T$ is minimal.  In Figure \ref{fig:1Dparabolic_fit_to_damped_cosine}(c) we
compare the numerically determined $\sigma^*$ with the approximations derived above.  The overall
shape is described qualitatively by any of the ODE approximations, but the midpoint approximation
\eqref{eq:damped_harmonic_oscillator_3} performs best quantitatively.

\subsection{Very small $k\eta$} \label{subsec_1D_verysmall}

The value of $\sigma^*$ calculated numerically for the algorithm becomes ill-defined when $k\eta$ is
very small, namely, when $k\eta\lesssim0.0027$ (Figure \ref{fig:introplot_1Dparabolic}).  The
analytic predictions, \eqref{eq:nu_eta_critical_1}, \eqref{eq:nu_eta_critical_2},
\eqref{eq:nu_eta_critical_3}, give negative answers for very small $k\eta$.  This indicates that, in
very shallow valleys or when a very small learning parameter is used, the benefit of
super-acceleration disappears.

At such small values of $k\eta$, any positive $\sigma$ is `above' the $\sigma^*$ curve and hence the
algorithm is automatically in the over-damped regime, so that the minimization dynamics is
non-oscillatory.  Moreover, since $\sigma$ appears in the algorithm only through the quantity
$\alpha=g-k\eta\sigma-1$, when $k\eta\sigma\ll(1-g)$ the super-acceleration parameter does not
affect the algorithm at all.  We will later see an example of this effect for a high-dimensional
system (Subsection \ref{sec:lregression}).

\subsection{Summary of one-parameter analysis}

In this Section, we provided details and derivations related to the message presented in the
Introduction (Figure \ref{fig:introplot_1Dparabolic}): that super-accelerating momentum-based
gradient descent is beneficial for minimization in the one-dimensional parabolic case.

Our analysis shows that either Nesterov-accelerated ($\sigma=g$) or super-accelerated ($\sigma>g$)
gradient descent relaxes to the minimum exponentially.  We can therefore associate a timescale $T$
to the relaxation dynamics.  Optimizing our algorithm means minimizing $T$.  With continuous-time
approximations, these algorithms are well-described as damped harmonic oscillators.  Since $T$
decreases with $\sigma$ in the underdamped regime and increases with $\sigma$ in the overdamped
regime, the $\sigma$ corresponding to \emph{critical} damping provides the minimal relaxation time.
Thus there is a well-defined optimal super-acceleration $\sigma^*$.  This was presented in the
Introduction (Figure \ref{fig:introplot_1Dparabolic}).

The result of the analysis is that $\sigma>1$ is preferable for all $k\eta$ values above a minimum.
(Numerically, we find this value to be $\approx0.0027$.)  This indicates that super-accelerating the
algorithm will give no benefit for minimization along very shallow valleys.  We will see examples of
this effect later.


\section{Higher-dimensional examples \label{sec:high_D}}

We now turn to the performance of super-acceleration in the minimization of multi-variable loss
functions.  In our analysis of the one-variable case with quadratic objective function, we found a
well-defined optimal value $\sigma^*$ of the super-acceleration parameter $\sigma$.  It is, of
course, impossible to generalize this prediction to an arbitrary multi-dimensional problem.  Even if
the dependences of $L(\theta)$ on each direction $\theta_i$ could be approximated as quadratic
($\approx \frac{1}{2}k_i\theta_i^2$), the constants $k_i$ corresponding to different directions can
be wildly different, making it impossible to identify an optimal super-acceleration $\sigma$.
Nevertheless, we expect that when the learning rate is not forced to be too small,
super-acceleration would be beneficial for many of the directions of $\theta$-space, so that it is
likely to be beneficial for the overall optimization.

In this Section, we test this idea on two synthetic multi-dimensional objective functions.
Subsection \ref{sec:2D} treats a non-quadratic function of two parameters.  In Subsection
\ref{sec:lregression} we treat quadratic $\approx50$-dimensional problems.  In both cases, we show
that using $\sigma>1$ provides benefits; at the same time we learn of some possible pitfalls.

\bigskip 

\subsection{Minimizing 2D function(s) \label{sec:2D}}

\begin{figure}[tb]
\begin{center}
\includegraphics[width=0.9\textwidth]{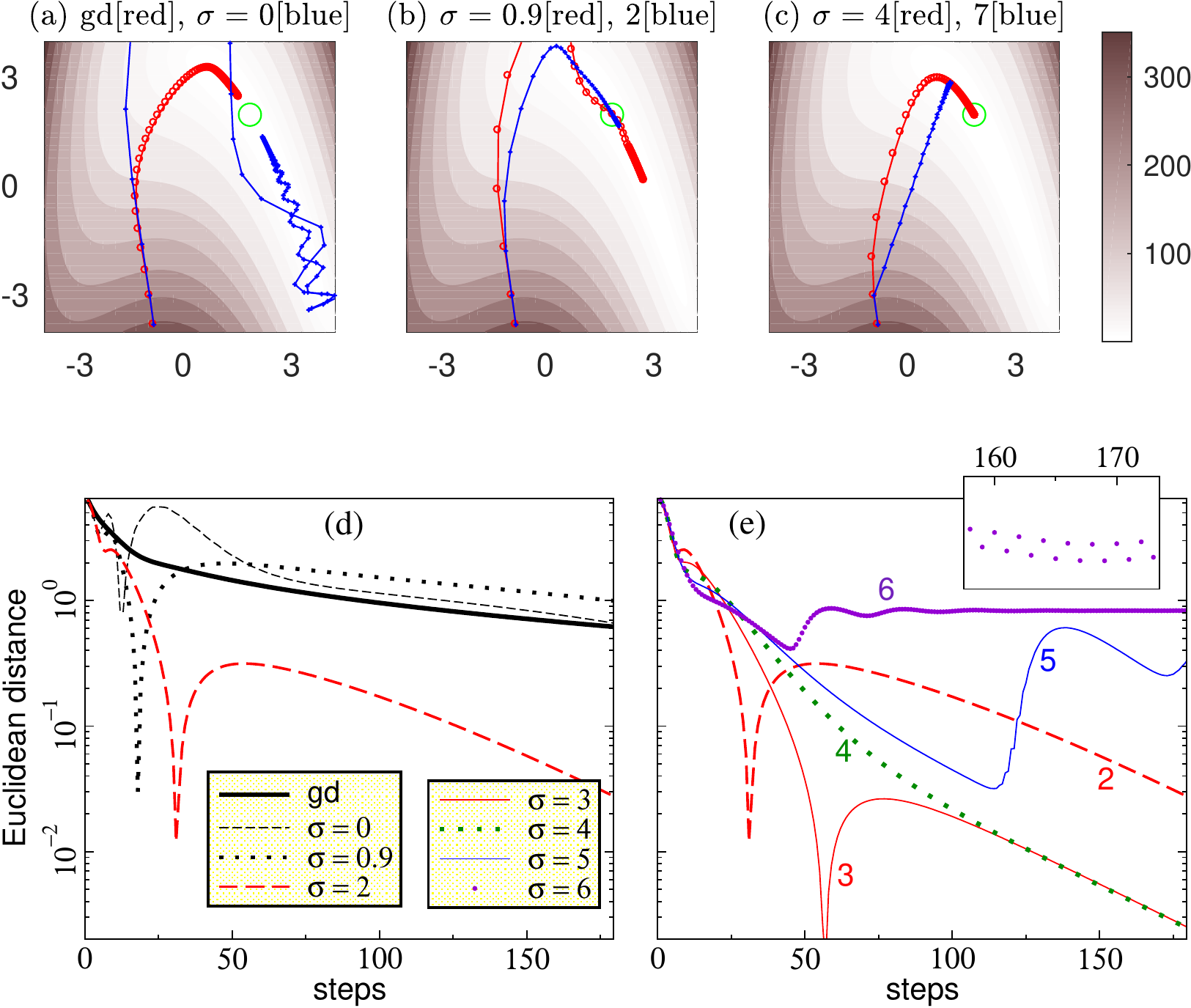}
\caption{Minimization of the objective function of Eq.\ \ref{eq:2D_functionA}.  First 180 steps,
  fixed $\eta=0.01$, starting point $(-1,-3.8)$. (a-c) Trajectories shown on $(\theta_1,\theta_2)$
  plane, together with contour lines of the function.  Green circle marks the minimum
  $\approx(1.690,1.963)$.  (a) Pure gradient descent (gd) and heavy ball
  ($\sigma=0$).  (b) $\sigma=0.9=g$ (Nesterov, red) and  $\sigma=
  2$ (blue).  (c) $\sigma= 4$ (red) and $7$ (blue).  Lower panels (d,e): evolution of the Euclidean
  distance to the minimum.  In (e), curves are marked with $\sigma$ 
  values for easy identification.   Inset to (e) zooms onto the $\sigma=6$ trajectory. 
\label{fig:2D_fnA_comboA}
}
\end{center}
\end{figure}

In Figure \ref{fig:2D_fnA_comboA} we show the effect of super-acceleration on gradient descent
minimization for the objective function
\begin{equation} \label{eq:2D_functionA}
L(\theta_1, \theta_2) = (\theta_1^2 + 2\theta_2 - 7)^2 + (2\theta_1 + \theta_2 - 5)^2 +  10^{-3}(\theta_1^6 + \theta_2^6)
\end{equation}
of two variables, $(\theta_1,\theta_2)$.  The selected function is  non-convex and its gradient is
nonlinear.  It is
designed to have a valley around which the `hills' have some spatial non-isotropic structure, as
seen in the contour plots in Figure \ref{fig:2D_fnA_comboA}.  The last term ensures that the
minimization procedure does not escape to infinity in any direction, and could be thought of as
mimicking a regularization term for machine learning.  In this Subsection, we consider  the effects of adding momentum and (super-)acceleration to gradient descent for this
potential, starting at the initial point $(-1,-3.8)$ and using a learning rate $\eta=0.01$.  We will
see that a super-acceleration value around $\sigma\approx4$ provides the best minimization.  Of course,
the exact performance and the benefits of super-acceleration will depend on the initial point and
the learning rate, even for this particular objective function.  However, we expect that there are
broadly applicable lessons to be learned from analyzing this particular case.

In Figure \ref{fig:2D_fnA_comboA} we show the trajectories of gradient descent with(out) momentum
and with various values of super-acceleration $\sigma$.  The first 180 steps are shown for each
minimization algorithm.  In the top row (a-c), trajectories are shown on the $\theta_1$-$\theta_2$
plane.  To avoid excessive clutter, each panel shows the trajectories of only two algorithms, one
with large empty red circles and one with small blue stars.  Successive positions are joined with
straight lines for better visualization of the trajectories.  In the bottom row, trajectories are
characterized using the Euclidean distance to the position of the minimum.

Panel \ref{fig:2D_fnA_comboA}(a) shows pure gradient descent without momentum, and also with
acceleration-less momentum, i.e., the heavy ball algorithm of Eq.~\eqref{eq:plain_momentum}
corresponding to $\sigma=0$.  Both these algorithms fall short of reaching the minimum in 180 steps.
The striking effect of adding momentum is that the trajectory performs wide oscillations and
overshoots.  It is often stated that adding momentum \emph{dampens} oscillation
\citep{Ruder_overview_2016, book_Goodfellow_Bengio_deeplearning_2016}, in apparent contradiction to
the observation in panel \ref{fig:2D_fnA_comboA}(a).  The situation in which momentum has a
dampening effect is when the learning rate is so large that pure gradient descent moves from one
side of a valley to another in each step, which appears to be the example(s) considered in
\citep{Ruder_overview_2016, book_Goodfellow_Bengio_deeplearning_2016}.  If $\eta$ is sufficiently
small, the situation observed in panel (a) is probably more typical: adding momentum causes strong
oscillations, but can speed up the minimization because the step sizes tend to be much larger.

In \ref{fig:2D_fnA_comboA}(b) we see that adding acceleration, $\sigma>0$, has a dampening effect on
the trajectory.  Also, super-acceleration ($\sigma=2$) works significantly better than usual
Nesterov acceleration ($\sigma=g=0.9$).  This improvement is also clearly seen in panel (d), where
the Eucledean distance to the minimum $(\theta_1^*,\theta_2^*)$ is plotted.  The distance falls much
faster with super-acceleration $\sigma=2$.  (In this particular case, adding momentum or Nesterov
acceleration does not improve the minimization compared to pure gradient descent, but this is not
generic --- typically momentum by itself will improve the algorithm.)

In \ref{fig:2D_fnA_comboA}(c) and \ref{fig:2D_fnA_comboA}(e) we show the effect of further
increasing $\sigma$.  For this case, the optimal values appear to be around $3$ or $4$.  As $\sigma$
is further increased, the algorithm tends to get trapped in regions other than the minimum.  For
example, \ref{fig:2D_fnA_comboA}(c) shows that the $\sigma=7$ algorithm shoots commendably fast into
the valley, but after that, gets stuck in a region which is not at the minimum.  We find that the
iteration oscillates between two nearby points.  In \ref{fig:2D_fnA_comboA}(e), inset, we show the
$\sigma=6$ algorithm approaching a similar situation.  It appears that the nonlinear map defined by
the super-accelearted iteration has attractors composed of points which are not at the minimum of
the loss function.  
The difference equations (map) describing the algorithm is nonlinear due to the nonlinearity of the
potential \eqref{eq:2D_functionA}.  In the algorithm \eqref{eq:super_Nesterov}, the $\sigma$
parameter appears inside the (gradient of the) loss function; hence it is not surprising that
nonlinear effects are enhanced when $\sigma$ is increased.  In the present case, when $\sigma$ is
increased too far, nonlinear effects are enhanced to the point of developing attractors at locations
which are not at the minimum.

Comparing in \ref{fig:2D_fnA_comboA}(e) the curves for $\sigma=3$ or $\sigma=4$, one notices that
$\sigma=3$ curve has a dip indicating that the algorithm overshoots the minimum and then returns
along the valley (as with smaller values of $\sigma$), but $\sigma=4$ does not have such an
overshooting behavior.  This can be understood from our previous one-parameter analysis applied to
dynamics within the valley: presumably, $\sigma\lesssim3$ corresponds to underdamping, and
$\sigma=4$ is closer to criticial damping.  This interpretation is meaningful only to the extent
that the valley may be approximated as a one-dimensional quadratic structure.

To summarize, we have found that the advantages of super-acceleration, i.e., increasing $\sigma$
beyond $g\approx1$, extends beyond the toy one-parameter problem analyzed in Section
\ref{sec:1D_quadratic}, and holds for non-quadratic potentials such that the gradient is a nonlinear
function of parameters.  However, in such a potential, for sufficiently large $\sigma$, the
nonlinearity may give rise to unwanted trapping effects.


\subsection{Linear regression \label{sec:lregression}}

We now consider applying gradient descent with super-acceleration to a
many-variable regression problem.

\begin{figure*}
\begin{center}
\includegraphics[width=1.0\textwidth]{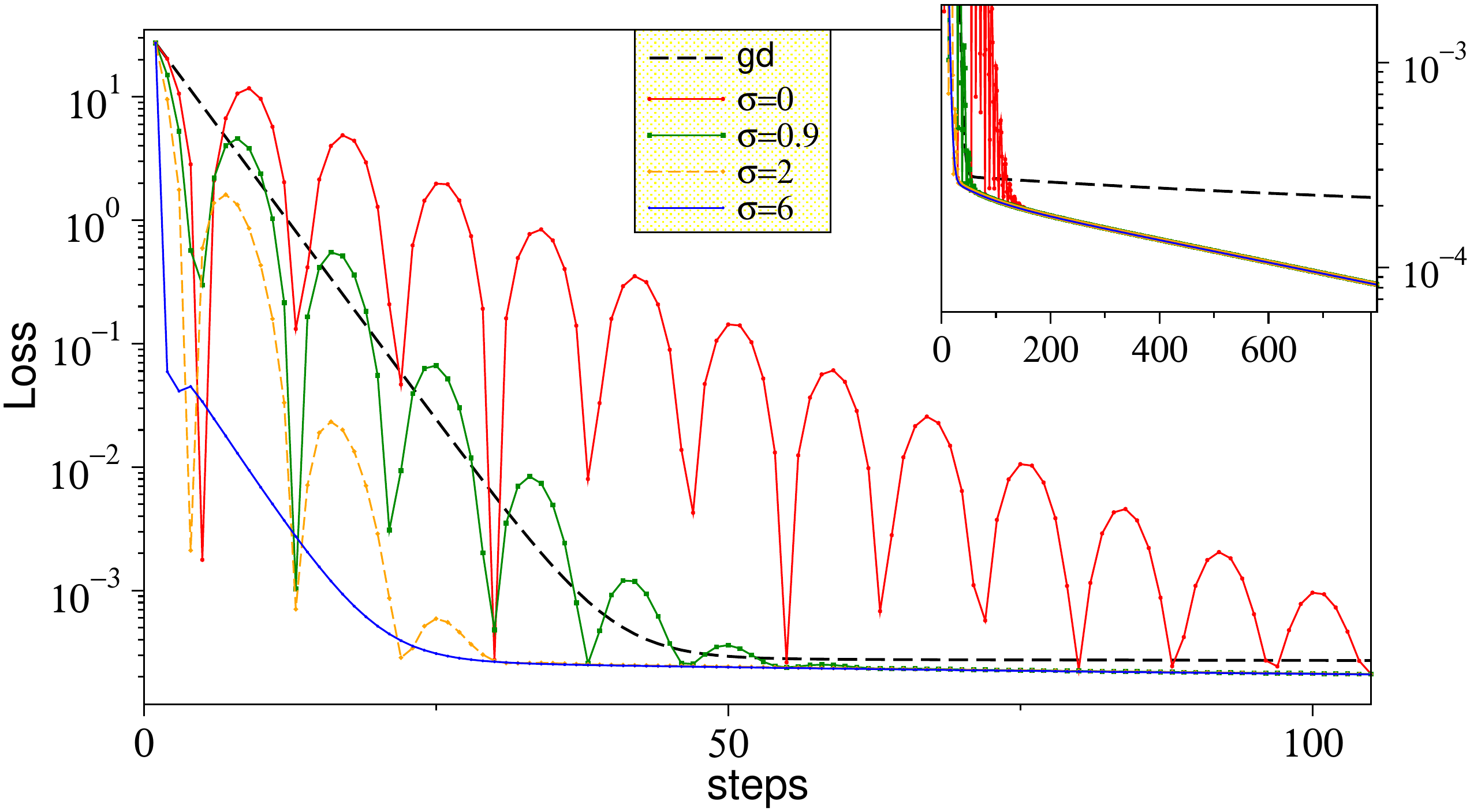}
\caption{The loss function for a 51-parameter linear regression model,
  evolving during gradient descent training with $\eta=0.005$.  
  Inset shows long-time behavior. 
\label{fig:linear_regression}
}
\end{center}
\end{figure*}

Consider real-valued observations $y$ which are functions of $N_{\rm f}$ real-valued features,
$x_\alpha$, with $\alpha=1$, $2$, $3$, \ldots, $N_{\rm f}$.  Given $N_{\rm data}$ such sets of
feature values and corresponding observations, the regression problem fits the data using a linear
model
\[
  h_{\theta}(x) = \theta_0+ \theta_1x_1+ \theta_2x_2+ \theta_3x_3+
  \ldots
  = \theta_0+ \sum_{\alpha=1}^{N_{\rm f}}\theta_{\alpha}x_{\alpha} . 
\]
The parameter space ($\theta$ space) is $(1+N_{\rm f})$-dimensional.
The loss function to be minimized in linear regression is 
\[
L(\theta) = \frac{1}{N_{\rm data}}\sum_{i=1}^{N_{\rm data}}  \left(  y^{(i)} - h_{\theta}(x^{(i)})
\right)^2
 = \frac{1}{N_{\rm data}} \sum_{i=1}^{N_{\rm data}}  
\left(  y^{(i)} - \theta_0 - \sum_{\alpha=1}^{N_{\rm f}} \theta_{\alpha}x_{\alpha}^{(i)}  \right)^2
. 
\]
Of course, linear regression can be solved by matrix methods which
may be more efficient than gradient descent.  Nevertheless, this
is a standard pedagogical setup for introducing gradient descent.  The
loss function for linear regression is known to be convex.  
Here we apply our algorithm, Eq.\ \eqref{eq:super_Nesterov}, to this
toy problem, to gain intuition about super-acceleration in
many-dimensional situations.   

In Figure \ref{fig:linear_regression} we present minimization algorithms for $N_{\rm f}=50$, i.e., a
$51$-dimensional problem.  We used $N_{\rm data}=1000$ data points, i.e., $N_{\rm data}=1000$ sets
of $({x_{\alpha}},y)$ values.  The $x_{\alpha}$ were chosen uniformly in $(0,1)$ and the $y$ values
were $y^{(i)} = \sum_{i} \left(x_{\alpha}^{(i)}/N_{\rm f}\right)^3$.  All $\theta_{\alpha}$ are set
to the same value $0.2$ at the start of each iteration.

Of course, the details are modified when other datasets, initialization, or $\eta$ are used.  We
have explored a number of variants, and observed that the aspects we emphasize below are broadly
applicable.

A curious aspect visible in Figure \ref{fig:linear_regression} (inset) is that, at late times in the
iteration, the iterations with momentum all follow the same path, irrespective of super-acceleration
$\sigma$.  Asymptotically, there appears to be a universal, $\sigma$-independent $L(\theta)$.  In
contrast, pure gradient descent without momentum trains slower (the loss function decreases slower)
than any of the momentum-based procedures.  The asymptotic curve is reached differently for
different $\sigma$, as seen in the main panel. As $\sigma$ is increased, the initial minimization is
more effective.  In fact, in this particular case, for initial minimization the $\sigma=0$ case is
slower even than pure gd, while the $\sigma=0.9$ case (Nesterov acceleration) is similar to pure gd.
The momentum-based methods perform better than pure gd only when super-acceleration,
$\sigma\gtrsim1$, is used.

These dynamical behaviors can be understood by decomposing the loss function into the
``eigen-directions'' of its hessian at the minimum.  Since the loss function is quadratic, it can
then be regarded as a sum of decoupled one-parameter quadratic objective functions, $L = \sum k_i
\gamma_i^2$, where the $\gamma_i$'s are linear combinations of the $\theta$ parameters.
The qualitative features of Figure \ref{fig:linear_regression} can be understood as a sum of the
decoupled cases of the one-parameter case treated in Section \ref{sec:1D_quadratic}.
The steeper directions (with larger $k_i$) are minimized faster by the algorithm.  For
$k_i\eta\gtrsim0.027$, we expect super-acceleration to be beneficial, this explains why larger
$\sigma$ provides faster initial minimization.  At late times, the directions with larger $k_i$ have
already been minimized, so that only the smallest $k_i$ value is relevant.  The dynamics is then in
the regime discussed in Subsection \ref{subsec_1D_verysmall}.  Any nonzero $\sigma$ puts us in the
overdamped region (above the $\sigma^*$ line of Figure \ref{fig:introplot_1Dparabolic}(b)); hence we
have the slow non-oscillatory decay seen at late times.  The timescale of this decay is nearly
independent of $\sigma$, as $\sigma$ appears in the theory only in the combination
$\alpha=g-1-k\eta\sigma$, which becomes effectively $\sigma$-independent when $k\eta\sigma\ll(1-g)$.

Given that the long-time behavior is identical for a range of $\sigma$, one might object that
super-acceleration provides no benefit in this case.  However, for practical machine-learning, the
initial rapid minimization is often the most relevant process, and the late-time fine minimization
might not even be desirable (e.g., might lead to overfitting).  For the initial minimization, Figure
\ref{fig:linear_regression} shows super-acceleration to be very beneficial.

\subsection{Lessons learned} 

From our studies with the two synthetic objective functions in this section, we have seen that
super-accelerating Nesterov momentum in general speeds up minimization, but that there could be
pitfalls.  In Subsection \ref{sec:2D}, we found that at large $\sigma$ nonlinearities might cause
convergence to spurious fixed points not located at the minimum.  In Subsection
\ref{sec:lregression} we saw a case where late-time minimization happens in an extremely shallow
direction, for which super-acceleration is no better (but no worse) than Nesterov acceleration or
even pure momentum.

\section{MNIST classification with neural networks \label{sec:NN_MNIST_classification}}




\begin{figure*}
\begin{center}
\includegraphics[width=1.0\textwidth]{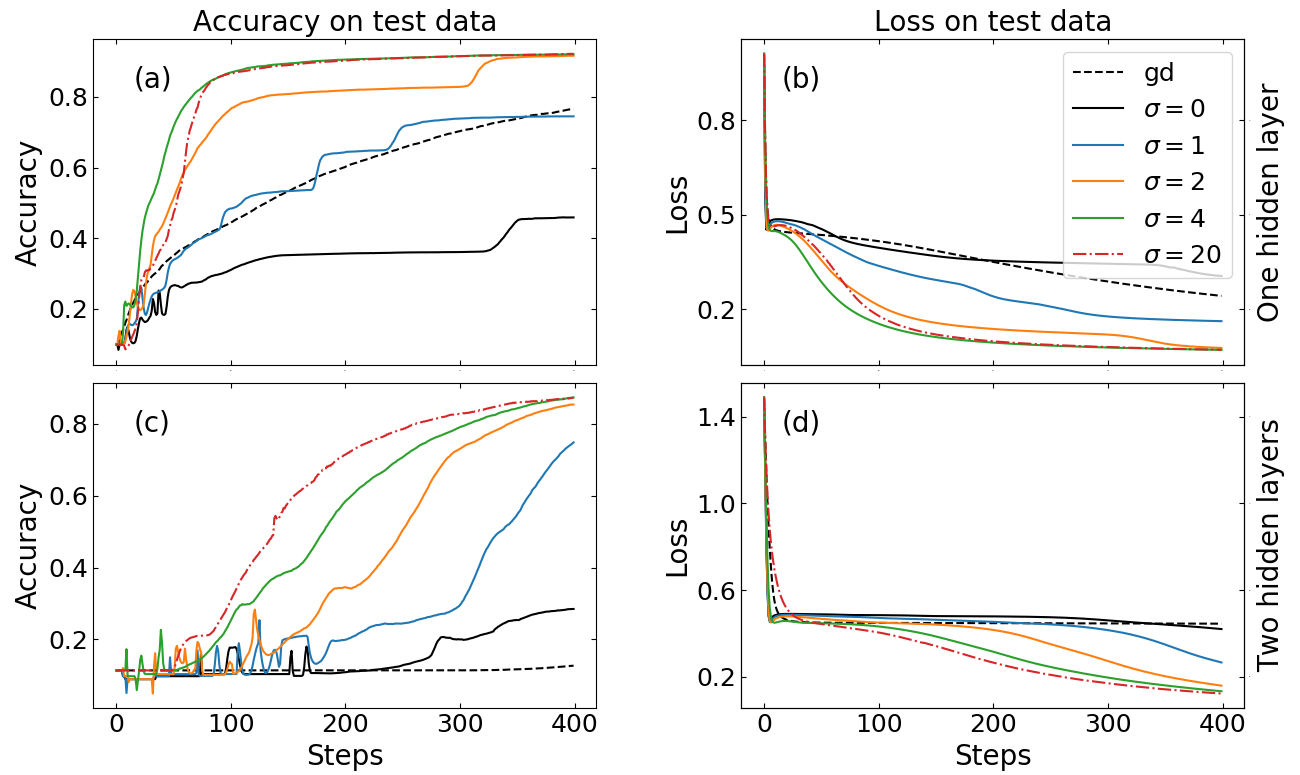}
\end{center}
\caption{Super-acceleration tested on training neural networks on
  MNIST digit classification.  The accuracy and loss are calculated on
  test data during training.  The gradient descent is non-stochastic:
  each step is an `epoch'.   (a,b) Single hidden layer: architecture
  $784$--$30$--$10$.
  (c,d) Two hidden layers:   $784$--$15$--$15$--$10$.
\label{fig:mnist_notstochastic}
} 
\end{figure*}


Having explored synthetic potentials and datasets in the previous section, we now turn to applying
super-acceleration to a standard machine learning benchmark problem: handwritten digit recognition
using neural networks, using the MNIST dataset \citep{LeCun_MNIST_1998,
  book_Nielsen_NeuralNetworks_2015}.  In Figure \ref{fig:mnist_notstochastic} we show results
obtained with gradient descent without stochasticity, i.e., instead of using mini-batches, all of
the training data is used for computing the gradient.  Thus each step is an epoch.

As usual, the training data consists of 60k images and the test data consists of 10k images.  Figure
\ref{fig:mnist_notstochastic} shows the accuracy and the loss function computed on the \emph{test}
data, as training progresses.  The input layer has $28^2=784$ nodes, each input being a pixel value.
The output layer has 10 nodes, representing desired classification results for the digits from 0 to
9.  The loss function used is the Euclidean distance of the output from the one-hot vector
corresponding to the digit label, averaged over the images.

To display the robustness of our results, we show the training process for two different network
architectures.  The panels on the top row, \ref{fig:mnist_notstochastic}(a,b), are for a network
with a single hidden layer with 30 nodes, i.e., a $784$--$30$--$10$ architecture.  The lower row shows
data for training a $784$--$15$--$15$--$10$ network, with two hidden layers.

We show both the classification accuracy and loss function on the test data.  (The curves for the
training data are very similar as there is little or no overfitting at these training stages.)  In
both cases, each training algorithm is started with the same values of parameters.  The weights and
biases are each drawn randomly from a gaussian distribution, centred at zero and with widths either
unity (biases) or normalized by the number of nodes in that layer (weights), as suggested in Chapter
3 of \citet{book_Nielsen_NeuralNetworks_2015}.  We run the minimization algorithms with learning
rate $\eta=0.5$ in each case.

Although some details are different for the two networks (top and bottom row of Figure
\ref{fig:mnist_notstochastic}), the outstanding feature in both cases is that super-acceleration
provides clear improvement in the training process.  In both networks, $\sigma=4$ provides superior
training compared to smaller values of $\sigma$.  We have also tested $\sigma=20$.  In the synthetic
examples of Section \ref{sec:high_D}, such large values generally led to instabilities or overflow
errors.  However, in the MNIST classification tasks, super-acceleration $\sigma=20$ apears to
function well, and even outperforms $\sigma=4$ in the two-layer case.  According to our
one-dimensional analysis in Section \ref{sec:1D_quadratic}, $\sigma=20$ would be beyond the optimal
value of $\sigma$, but in the MNIST case large $\sigma$ values apparently continue to be performant.

Similar to the multi-dimensional quadratic case studied in Subsection \ref{sec:lregression}, we note
that, late in the training process, the curves with momentum appear to merge together independent of
$\sigma$.  In our data up to 400 steps, the curves for $\sigma\geq2$ have already merged.  In
analogy to Figure \ref{fig:linear_regression}, one surmises that the other curves will merge later
on during training.  This suggests that the loss landscape for MNIST classification has some
similarities to that of the regression problem, i.e., to a multi-dimensional quadratic function with
a range of curvatures in different directions.

\begin{figure*}
\begin{center}
\includegraphics[width=1.0\textwidth]{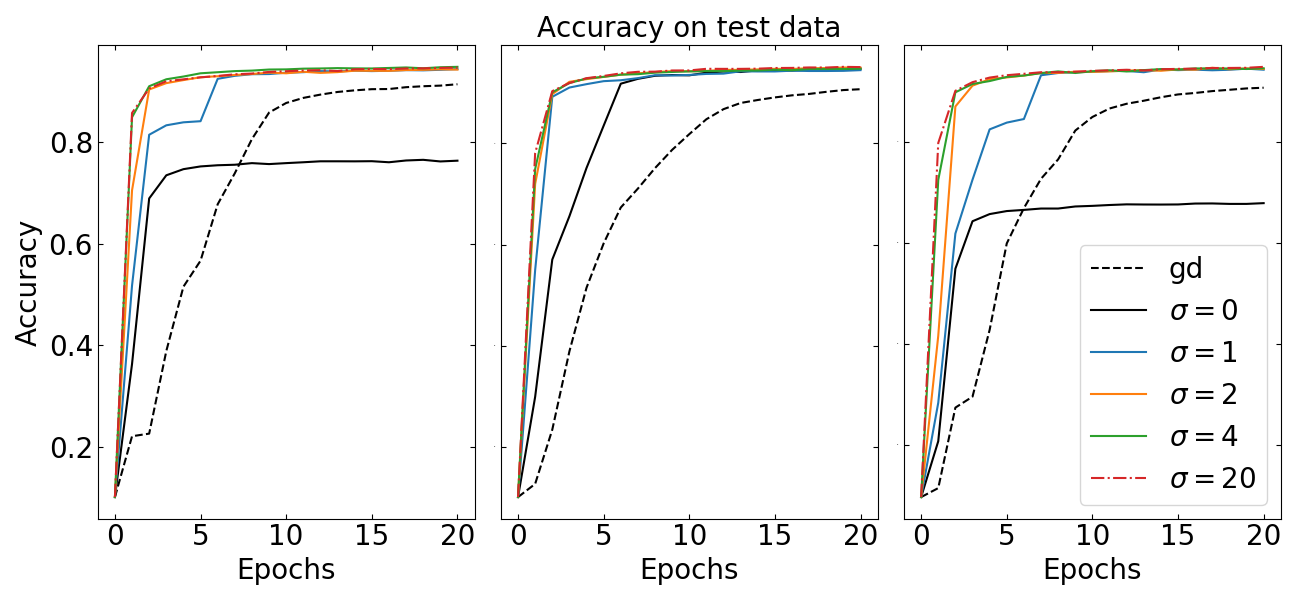}
\end{center}
\caption{Super-acceleration tested on training neural networks on
  MNIST digit classification, with the  $784$--$30$--$10$ network described for the previous figure.
  Now, stochastic gradient descent is used.  The training set is divided into mini-batches, each
  containing $200$ training images.  Each epoch now consists of $60000/200 = 300$ steps.  Three
  different stochastic runs are shown. 
\label{fig:mnist_stochastic}
}
\end{figure*}

In Figure \ref{fig:mnist_stochastic}, we show evaluations of super-acceleration using
\emph{stochastic} gradient descent, which is more common in practical machine learning.  The
training set is now stochastically divided into mini-batches, and the mini-batches are used to
calculate approximations to the gradient.  The overall features are similar to the non-stochastic
case of Figure \ref{fig:mnist_notstochastic}:\\ $[a]\;$ Larger $\sigma$ allows rapid optimization;
hence super-acceleration is beneficial.\\ $[b]\;$ In the long-time limit, the accuracy or loss tends
to become independent of $\sigma$.


\section{Combining with adaptive methods (RMSProp)}  \label{sec:RMSPROP}

The learning rate $\eta$ is a hyperparameter that obviously affects the minimization process
strongly.  Therefore, it is tempting to modify the effective learning rate based on data acquired
during the process.  In recent years, many such adaptive algorithms have appeared, and at this time
these algorithms have arguably become more common than gradient descent with constant or
pre-determined $\eta$.  In particular, RMSProp \citep{rmsprop_Courseralecture_2012} and Adam
\citep{Kingma_Ba_adam_2014} are often mentioned as default minimization algorithms.

When RMSProp is combined with momentum, it is almost identical to Adam
\citep{Ruder_overview_2016, book_Goodfellow_Bengio_deeplearning_2016}.
Nesterov-accelerating these methods, by evaluating the gradient at the
estimated one-step-forward position, have also been discussed
\citep{Dozat_Nadam_2016, Ruder_overview_2016}.  Clearly, if the
algorithm can be Nesterov accelerated, it can also be
super-accelerated, in the sense introduced in the present work.  A
thorough understanding of the effects and potential benefits of
combining adaptive learning rates with super-acceleration would
require a very extensive study; here we present some basic
calculations.  

The RMSProp algorithm with momentum and super-acceleration is 
\begin{equation}  \label{eq:super_RMS_general}
 \begin{split}  
 r^{(i)} &= \beta_2 r^{(i-1)} + (1-\beta_2)  \left[\nabla L(\theta^{(i)} + \sigma m^{(i-1)})\right]^2 \\
m^{(i)} &= gm^{(i-1)} - \frac{\eta}{\sqrt{r^{(i)} + \epsilon}} \nabla L(\theta^{(i)} + \sigma m^{(i-1)}) \\
\theta^{(i+1)} &= \theta^{(i)} + m^{(i)}  . 
\end{split}
\end{equation}
We have taken the RMSProp+momentum algorithm as described in \citep{Ruder_overview_2016,
  book_Goodfellow_Bengio_deeplearning_2016}, and added super-acceleration, i.e., the gradient is now
evaluated not at the current values of the parameters, but at an estimated forward point
$\theta^{(i)} + \sigma m^{(i-1)}$.  The values $\beta_2=0.999$ and $\epsilon=10^{-7}$ are reasonably
standard; we confine ourselves to these values.  The `division' in the second line is component-wise
division, which is not standard mathematical notation but standard in this field.  The intuition
behind this algorithm is that directions in $\theta$ space which have seen little change (smaller
gradient magnitudes) in recent iterations are favored at the expense of those directions that have
enjoyed rapid changes (larger gradient magnitudes).  This is supposed to help escape from saddle
points and minimize rapidly along shallow directions.

The Adam and Nadam algorithms are based on almost the same ideas \citep{Kingma_Ba_adam_2014,
  Dozat_Nadam_2016, Ruder_overview_2016, book_Goodfellow_Bengio_deeplearning_2016}.  The main
differences are (1) that the adaptive factor appears in the third equation for $\theta$ rather than
in the second equation for $m$; (2) a bias correction that is important for initial iterations,
which we believe does not play much of a role in determining the timescale.  We do not currently
have a good theoretical understanding of the differences between RMSProp (with momentum) and Adam.
It is reasonable to expect that the basic intuitions obtained below for combining
super-accelearation with RMSProp should also hold for combining super-acceleration with other
adaptive algorithms.

We provide below some results for the algorithm
\eqref{eq:super_RMS_general} for the one-parameter case with a
parabolic loss function (Subsection \ref{subsec:RMSPROP_1Dparabolic})
and with the two-parameter loss function, Eq.\ \eqref{eq:2D_functionA}, used
previously (Subsection \ref{subsec:RMSPROP_2DfnA}).  We reiterate 
that this should be a regarded as an initial overview and a full
analysis exploring various parameter regimes would involve a
substantial study in its own right.

\subsection{One-parameter parabolic function}
\label{subsec:RMSPROP_1Dparabolic}

\begin{figure*}
\begin{center}
\includegraphics[width=1.0\textwidth]{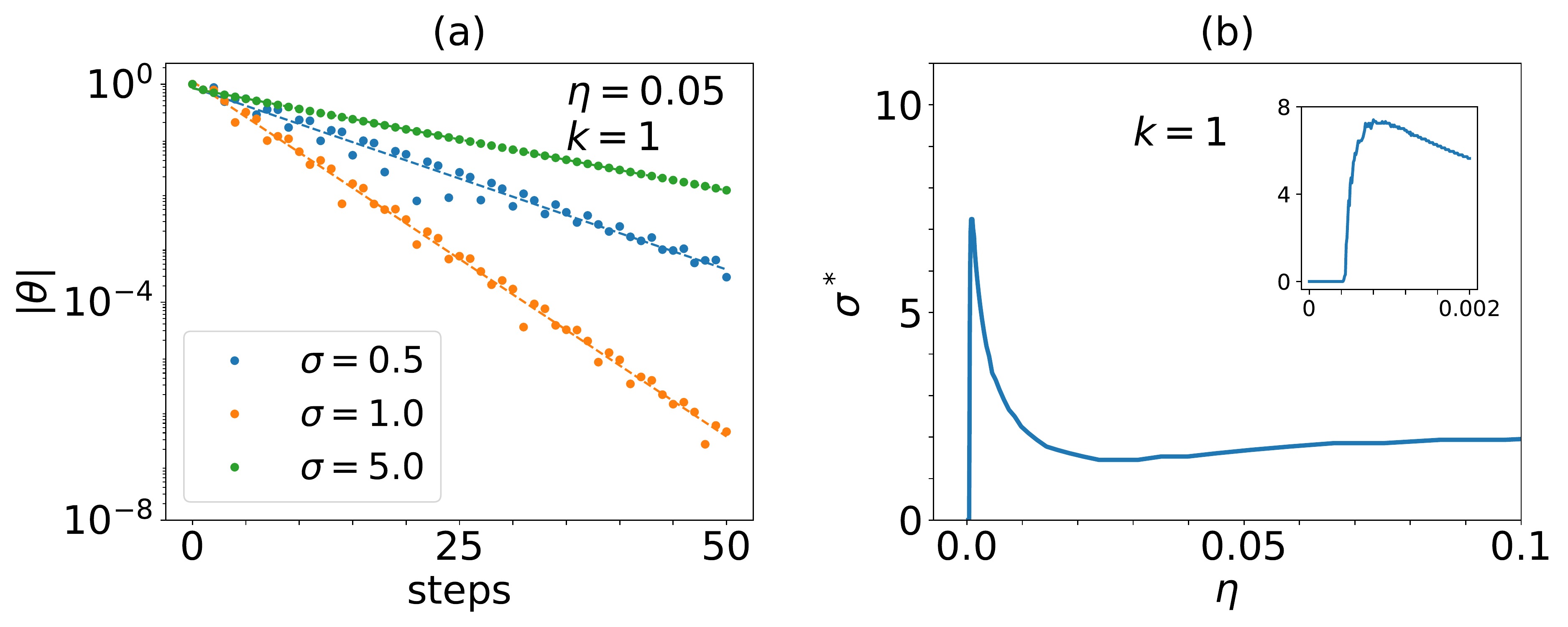}
\caption{Super-accelerating RMSProp with momentum: parabolic objective
function for a single $\theta$ parameter.  Hyperparameters:
$\beta_2=0.999$ and
$\epsilon=10^{-7}$. (a)  $|\theta^{(i)}|$ vs $i$, fitted to
exponentials, in order to extract `timescale' $T$.  (b) Optimal
super-acceleration as a function of learning rate.  
\label{fig:RMPSProp_1Dparabolic}
}
\end{center}
\end{figure*}

As we did in Section \ref{sec:1D_quadratic} for the non-adaptive super-acceleration algorithm, we
now examine RMSProp with super-acceleration for the simplest model: a loss function of the form
$L(\theta)= \frac{1}{2}k\theta^2$, where $\theta$ is a single real variable, and \-$k$ is a positive
constant. The algorithm \eqref{eq:super_RMS_general} for this special case becomes
\begin{equation}  \label{eq:super_RMS_1D}
 \begin{split}  
 r^{(i)} &= \beta_2 r^{(i-1)} + (1-\beta_2) k^2 \left[\theta^{(i)} + \sigma m^{(i-1)}\right]^2 \\
m^{(i)} &= gm^{(i-1)} - \frac{k\eta}{\sqrt{r^{(i)} + \epsilon}} \left[ \theta^{(i)} + \sigma m^{(i-1)}\right] \\
\theta^{(i+1)} &= \theta^{(i)} + m^{(i)}  . 
\end{split}
\end{equation}
with the understanding that $\theta$, $m$ and $r$ now have single-component real values instead of
having many components. (If they are regarded as vectors in $\mathbb{R}^D$, then $D=1$.)  If
Eq.\ \eqref{eq:super_RMS_1D} is regarded as a map of the variable $\theta$, then it is clearly a
heavily nonlinear map. As in Section \ref{sec:1D_quadratic}, we can find ordinary differential
equations approximating this discrete algorithm; these ODE's are nonlinear and therefore not as easy
to draw lessons from as was the case with the damped harmonic oscillators.  Instead, we present in
Figure \ref{fig:RMPSProp_1Dparabolic} results from direct analysis of numerical runs of the
algorithm \eqref{eq:super_RMS_1D}.

Unlike the non-adaptive case in Section \ref{sec:1D_quadratic}, in Eq.\ \eqref{eq:super_RMS_1D} the
constant $k$, parameterizing the steepness of the parabolic potential, cannot be absorbed into a
factor $k\eta$.  It is thus an independent parameter, meaning that an exploration of $k$ values is
necessary for a full understanding of the algorithm.  For the purpose of the present preliminary
exploration, we restrict to $k=1$.

Although the damped harmonic oscillator \eqref{eq:damped_harmonic_oscillator_0} is no longer an
excellent approximation, we find that the algorithm has the same overall qualitative features as in
Section \ref{sec:1D_quadratic}: for small $\sigma$ we obtain oscillatory relaxation to the minimum
$\theta=0$, and the relaxation time scale $T$ decreases with increasing $\sigma$, while above a
`critical' $\sigma$ the relaxation is non-oscillatory and increases with further increase of
$\sigma$.  Thus there is an optimal value of $\sigma$ which minimizes $T$.  Despite the qualitative
similarity, a function of the form $e^{-t/T}\cos(\bar{\omega}t+\delta)$ does not provide
quantitative fits to the $\theta^{(i)}$ vs $i$ data (identfying $i$ with time as before).
Therefore, we simply fit the function $\propto e^{-t/T}$ to $\left|\theta^{(i)}\right|$, in order to
extract $T$, effectively averaging over the oscillations, Figure \ref{fig:RMPSProp_1Dparabolic}(a).
This procedure is not theoretically rigorous but suffices to extract a relaxation time scale.  As
before, plotting $T$ against $\sigma$ provides us with the optimal super-acceleration value
$\sigma^*$.  Figure \ref{fig:RMPSProp_1Dparabolic}(b) shows $\sigma^*$ as a function of $\eta$ for
$k=1$.

From Figure \ref{fig:RMPSProp_1Dparabolic}(b) we conclude that a
super-acceleration parameter $\sigma$ considerably larger than
$g\approx1$ is beneficial, as long as the learning parameter $\eta$ is
not very small.  This is very similar to the non-adaptive case.

\subsection{Two-dimensional function} 
\label{subsec:RMSPROP_2DfnA}

\begin{figure*}
\begin{center}
\includegraphics[width=1.0\textwidth]{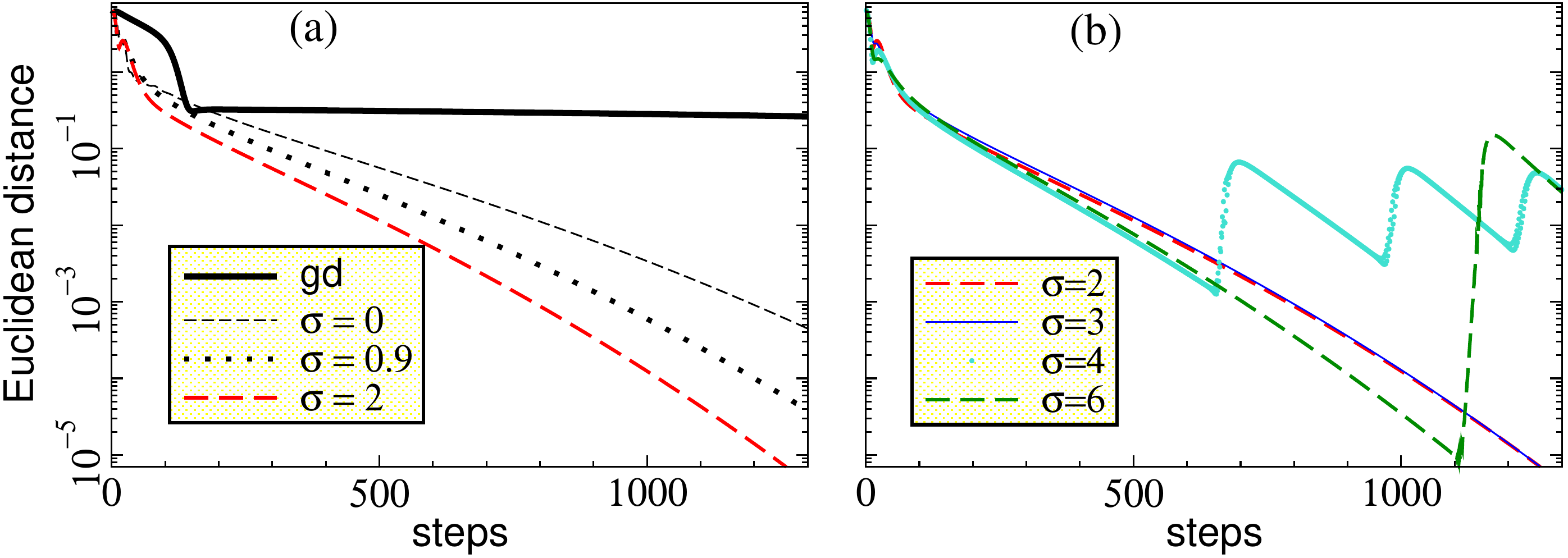}
\caption{Minimization of the objective function of Eq.\
  \ref{eq:2D_functionA}, using RMSProp with momentum and
  (super-)acceleration, shown using the Euclidean distance from the
  minimum.   As in Figure \ref{fig:2D_fnA_comboA},
  $\eta=0.01$ and the starting point is
  $(\theta_1,\theta_2)= (-1,-3.8)$.  RMSProp hyperparameters are
  $\beta_2=0.999$ and $\epsilon=10^{-7}$.
\label{fig:2D_fnA_RMSProp}
}
\end{center}
\end{figure*}

We have performed some exploratory calculations of  RMSProp with
super-acceleration, algorithm \eqref{eq:super_RMS_general}, for
two-dimensional test functions.  Overall, the situation is less clear
than the non-adaptive case, and the algorithm seems more likely to
show classic behaviors associated with nonlinearlities, such as
attractions to regions other than the minimum.  However, the general 
trend is still that super-acceleration with $\sigma>1$ is beneficial.  

In Figure \ref{fig:2D_fnA_RMSProp}, we show the minimization process for the two-parameter function,
Eq.\ \ref{eq:2D_functionA}, explored in Subsection \ref{sec:2D}, for $\eta=0.1$ and the same
starting point as in Figure \ref{fig:2D_fnA_comboA}.  As in the non-adaptive case, the minimization
becomes faster as $\sigma$ is increased beyond $g\approx1$, but when $\sigma$ is too large,
nonlinearities cause the algorithm to converge elsewhere instead of at the minimum.
These results suggest that it might be generally advantageous to start with a large $\sigma$ for
rapid initial minimization, and then at later times reduce $\sigma$ so as to avoid nonlinearities.


\section{Discussion, Context and Recommendations} \label{sec:concl}

%
Currently, there is no consensus on the `best' variant of gradient descent for use in machine
learning, and minimization methods continue to be an active research topic.
The most basic improvements of gradient descent are momentum and Nesterov acceleration. 
There is a large body of current research either analyzing or suggesting modifications to
(non-adaptive) momentum-based methods \citep{ Wibisono_Wilson_accelerated_methods_2015,
  Wilson_Recht_Jordan_lyapunov_analysis_2016, Wibisono_Wilson_Jordan_variational_PNAS2016,
  YuanYing_influence_of_acceleration_2016, Jin_Jordan_accelGD_escape_2017,
  Lucas_aggregated_momentum_2018, Ma_Yarats_quasihyperbolic_2018, Cyrus_Hu_Lessard_robust_2018,
  Srinivasan_Sankar_ADINE_2018, 
  Kovachki_Stuart_analysis_momentum_2019, Chen_Kyrillidis_decaying_momentum_2019,
  Gitman_Lang_role_of_momentum_2019}.

The present work can be considered a contribution of this type.  We have proposed a simple twist on
the idea of accelerated momentum --- extending the acceleration idea of `looking ahead' to beyond a
single estimated step.  We have also presented an analysis of the momentum and Nesterov acceleration
algorithms in terms of continuous-time ordinary differential equations (ODEs), and exploited the
interpretation of these ODE's as damped harmonic motion to predict the optimal super-acceleration
parameter, $\sigma$.  The idea that moving toward the critical point is advantageous has appeared
before in the literature \citep{Ning_momentum_1999}.  We have used this idea quantitatively to
predict the optimal $\sigma$.

Although adaptive methods (RMSProp, Adam, etc.) are generally considered to be better performant,
the relative merits of adaptive versus non-adaptive methods are not yet fully understood
\citep{Wilson_Recht_marginal_adaptive_2017, Reddi_etal_convergence_of_Adam_2019,
  Choi_Schallue_empirical_comparisons_2019}.
Naturally, there is also a significant amount of current research on analyzing and improving
adaptive methods \citep{De_Mukherjee_convergence_RMSProp_ADAM_Nesterov_2018,
  Zhou_padam_convergence_2018, Ma_Yarats_quasihyperbolic_2018, Barakat_Bianchi_Adam_ODE_2018, 
  Luo_Xiong_Sun_adaptive_dynamicbound_2019, Tong_Liang_Bi_calibrating_adaptive_2019,
  Wu_Du_Ward_Adaptive_Overparametrized_2019, Zhang_Ba_Hinton_lookahead_2019}.
In the present work, we have presented some preliminary results on how our super-acceleration ideas can
also be incorporated into adaptive methods such as RMSProp or Adam/Nadam.  

We have described a variety of cases where super-acceleration ($\sigma>1$) speeds up optimization.
Our exploration has also revealed two ways in which super-acceleration may fail to provide an
advantage.  First, in a very shallow part of the landscape, the algorithm may be in the regime of
ulra-small $k\eta$, in the language of Section \ref{sec:1D_quadratic}.  In such a case, the
algorithm becomes nearly $\sigma$-independent.  Another possible pitfall is due to large
nonlinearities, which can be amplified by $\sigma$ to the extent that the algorithm can get trapped
in a non-extremal region.  Examples were seen in Subsections \ref{sec:2D} and
\ref{subsec:RMSPROP_2DfnA}.  For the severely non-quadratic synthetic function of
Eq.\ \eqref{eq:2D_functionA}, such instabilities or attractors appeared for $\sigma\gtrsim4$.  Note
however that we did not see this type of phenomenon in our experimentation with neural networks
(MNIST classification, Section \ref{sec:NN_MNIST_classification}), even for much larger $\sigma$
values, which suggests that these problems (nonlinear instabilities) might not be serious in
practical applications.

In view of these considerations, we can provide the following recommendation for using gradient
descent with super-acceleration: start the algorithm with a relatively large $\sigma$, say around 5,
and then gradually lower $\sigma$ to around $2$ so as to avoid spurious nonlinear phenomena.

This work opens up some questions which require further investigation.  (1)~First, a thorough
analysis and exploration of super-acceleration in combination with the standard adaptive methods ---
RMSProp, Adam and Adadelta --- remains to be performed.  In Section \ref{sec:RMSPROP} we have
provided a very preliminary look.  Our initial results are promising and suggest that
super-accleration might improve these algorithms significantly.  In particular, it is possible that
in combination with Adadelta, super-acceleration might not suffer from the small $k\eta$ problem, as
the effective learning rate $\eta$ might be increased when the slope (effective $k$) becomes too
small.  (2)~Second, we have focused on non-stochastic minimization, except for a cursory exploration
in Section \ref{sec:NN_MNIST_classification} in the context of MNIST digit classification.  A
thorough analysis of super-acceleration for stochastic gradient descent remains another task for
future work.


\bibliographystyle{plainnat}
\bibliography{refs_GradDesc}

\end{document}